# A Non-Numeric Approach to Multi-Criteria/Multi-Expert Aggregation Based on Approximate Reasoning

Ronald R. Yager
Machine Intelligence Institute
Iona College
New Rochelle, NY 10801

## Abstract

We describe a technique that can be used for the fusion of multiple sources of information as well as for the evaluation and selection of alternatives under multi-criteria. Three important properties contribute to the uniqueness of the technique introduced. The first is the ability to do all necessary operations and aggregations with information that is of a nonnumeric linguistic nature. This facility greatly reduces the burden on the providers of information, the experts. A second characterizing feature is the ability assign, again linguistically, differing importances to the criteria or in the case of information fusion to the individual sources of information. A third significant feature of the approach is its ability to be used as method to find a consensus of the opinion of multiple experts on the issue of concern. The techniques used in this approach are base on ideas developed from the theory of approximate reasoning. We illustrate the approach with a problem of project selection.

## 1. Introduction

A problem of considerable interest is the so called *information fusion problem*. In this problem one has a number sources of information bearing on a set of hypothesis. The objective here is to aggregate these different sources of information to get some indication of the validity of the individual hypothesis. Pattern recognition can be seen as a special case of this problem. The difficulty of this problem can be somewhat compounded if we allow multiple experts to participate in the validation of hypothesis based upon source information. This becomes even more complicated if each expert can have a different interpretation of the being of a source of information on the validity of a hypothesis as well as attributing a different importance to each of the sources of information. Medical diagnosis is a classic example of this environment. A problem with the exact same structure as the above is the multi-criteria decision problem. In this environment the set of hypothesis are replaced by a set decision alternatives.. The sources of information are replaced by criteria which must be satisfied by good solutions. Again one can also consider an environment in which a number of experts participate in the actual selection process, a kind of consensus, but where each expert has a different degree of importance associated with each criteria. Generically we shall call this class of problems **multi-criteria/ multi-expert aggregation.** We shall here present a procedure for the solution of this problem. The procedure presented here satisfies the added restriction that the information provided by the experts need only be of a linguistic/non-numeric nature. The ability to aggregate nonnumeric information greatly reduces the burden imposed on the experts in providing their information as well as freeing us from the so called tyranny of numbers.

The procedure to be described combines an approach suggested by Yager (1981), which was recently discussed by Caudell (1990), for aggregating fuzzy sets and Yager's recent work on linguistic quantifiers using ordered weighted averaging (OWA) operators (Yager 1988, To Appear). The spirit of this procedure can be said to a formalization of a kind approximate reasoning.

In order more tangible the ideas discussed we of shall consider the description of the technique in the framework of the problem of selecting from a set of alternative project proposals those which are to be funded. Essentially the funding selection process can be seen as a multi-criteria decision process. Central to any multi-criteria decision process is the necessity to aggregate criteria satisfaction. A requirement for aggregation is that the information provided must be on a scale of sufficient sophistication to allow appropriate aggregation operations to be performed. One such scale having this property is the numeric scale. One problem with such a numeric scale is that we become subject to an effect which I shall call *the tyranny of numbers*. The essential issue here is that the numbers take a life and precision far in excess of the ability of the evaluators in providing these scores. The approach we suggest allows for the aggregation of multi-criteria but avoids the tyranny of numbers by using a scale that essentially only requires a linear ordering.

The procedure described here can be seen as a two stage process. In the first stage, individual experts are asked to provide an evaluation of the alternatives, the different proposals. This evaluation consists of providing a measure of how well each of the criteria required of a good solution are satisfied by each of the alternatives. In addition each expert provides an indication of how important he thinks each criteria is. The values to be used for the evaluation of the ratings of the alternatives and importances will be drawn from a linguistic scale making it easier for the evaluator to provide the information. We use a methodology which we developed in (Yager 1981) to provide, for each expert, an overall rating of each alternative for that experts inputted information. In the second stage, we use a methodology, based upon OWA operators, which we introduced in (Yager 1988) and extended in (Yager To Appear) to aggregate the individual experts evaluations to obtain a combined overall rating for each object. This overall evaluation can then be used by the decision maker as an aid in the selection process.

In the application specifically addressed in this paper, we augment the ratings by some textual evaluation. While the need for textual evaluation is not required for the implementation of the procedure described, we feel that it can provide useful additional information in the selection process.

## 2. PROBLEM FORMULATION

The problem we are interested in addressing can be seen to consist of four components. The first component is a collection $P = \{P_1, \ldots P_n\}$ of proposals from amongst which we desire to select some to be funded. In the information fusion problem this would be our collection of hypothesis.

The second component is a group of experts whose advise is solicited in helping make the decision, we denote this as $A = \{A_1, \ldots A_q\}$. Generally q is much smaller than n.

The third component of the system are the set of criteria to be used in evaluating the suitability of the alternatives. In the information fusion problem instead of criteria we would have the information provided by the different sources of information.

The fourth component of the system is the decision maker who has executive responsibility for using(aggregating) the advise of the experts and then making the final decision.

In this environment, the function of the advisory panel of experts is to provide information to help the decision maker. The following procedure can be used to accomplish this. Each expert is given information about each project and is asked to fill out a form regarding these projects. The form consists of questions relating to that proposal, these questions constitute the criteria. Each question requires a two part answer. The first part consists of a linguistic score drawn from a scale provided and the second part consists of a textual portion. The purpose of the textual portion is to help amplify and clarify the score. The scores play central role in this process. At the very least (and very best) they help make the broad distinction between those proposals that are very bad and those that are very good. They also help provide some ordering amongst the proposals.

It should be noted that in the information fusion problem each of the experts would be provided with the observations from the different sources, instead of the questions or criteria. In addition each would be asked to indicate the degree to which the source supports each of the hypothesis under consideration.

## 3. A Non-Numeric Technique Multi-Criteria Aggregation

In this section, we shall assume each questionaire, consists of n items. As noted in the previous section each item consists of a criteria of concern in evaluating a proposal. Each expert will select a value from the following scale S:

| Perfect (P) | $S_7$ |
| Very High (VH) | $S_6$ |
| High (H) | $S_5$ |
| Medium (M) | $S_4$ |
| Low | $S_3$ |
| Very Low | $S_2$ |
| None | $S_1$ |

to indicate the degree to which a proposed project satisfies a criteria.

The use of such a scale provides of course a natural ordering, $S_i > S_j$ if $i > j$. Of primary significance is that the use of such a scale doesn't impose undue burden upon the evaluator in that it doesn't impose the meaningless precision of numbers. The scale is essentially a linear ordering and just implies that one score is better then another. However, the use of linguistic terms associated with these scores makes it easier for the evaluator to manipulate. The use of such a seven point scale appears also to be in line with Miller's (1969) observation that human beings can reasonably manage to keep in mind seven or so items.

Implicit in this scale are two operators, the maximum and minimum of any two scores:

$$Max(S_i, S_j) = S_i \quad \text{if } S_i \geq S_j$$
$$Min(S_i, S_j) = S_j \quad \text{if } S_j \leq S_i$$

We shall denote the max by $\vee$ and the min by $\wedge$.

Thus for any arbitrary proposal $P_i$ each expert will provide a collection of n values.

$$(P_{ik}(q_1), P_{ik}(q_2), \ldots P_{ik}(q_n))$$

where $P_{ik}(q_j)$ is the rating of the $i^{th}$ proposal on the $j^{th}$ criteria by the $k^{th}$ expert. Each $P_{ik}(q_j)$ is an element in the set S of allowable scores.

Assuming n = 6, a typical scoring for proposal from one expert would be:

$P_{ik}$: (high, medium, low, perfect, very high, perfect)

Independent of this evaluation procedure each criteria must be given a measure of importance. Two methods are allowable for the determination of importances. In the first approach the importances are assigned by the experts themselves. In the second approach the executive decision maker assigns a measure of importance to each of the criteria. This information may or may not be available to the evaluators, it is not required by the evaluator.

In either approach the above scale is used to provide the importance associated with the criteria. It should be noted that the only requirement on the assignment of importances is that the most important criteria is given the rating P. We shall use $I(q_j)$ to indicate the importance associated with the criteria. A possible realization for importances could be

$$I(q_1) = P$$
$$I(q_2) = VH$$
$$I(q_3) = VH$$
$$I(q_4) = M$$
$$I(q_5) = L$$
$$I(q_6) = L$$

The next step in the process is to find the overall valuation for a proposal by a given expert.

In order to accomplish this overall evaluation, we use a methodology suggested by Yager (1981). This approach was recently discussed by Caudell (1990).

A crucial aspect of this approach is the taking of the negation of the importances. Yager (1981) introduced a technique for taking the negation on a linear scale of the type we have used. In particular, he suggested that if we have a scale of q items of the kind we are using then

$$Neg(S_i) = S_{q-i+1}$$

We not that this operation satisfies the desirable properties of such a negation as discussed by Dubois & Prade (1985).

(1) Closure
    For any $s \in S$, $Neg(s) \in S$
(2) Order Reversal
    For $S_i > S_j$, $Neg(S_i) \leq Neg(S_j)$
(3) Involution
    $Neg(Neg(S_i)) = S_i$    for all i

For the scale that we are using, we see that the negation operation provides the following

Neg(P) = N         $(Neg(S_7) = S_1)$
Neg(VH) = VL       $(Neg(S_6) = S_2)$
Neg(H) = L         $(Neg(S_5) = S_3)$
Neg(M) = M         $(Neg(S_4) = S_4)$
Neg(L) = H         $(Neg(S_3) = S_5)$
Neg(VL) = VH       $(Neg(S_2) = S_6)$
Neg(N) = P         $(Neg(S_1) = S_7)$

The methodology suggested by Yager (1981) which can be used to find the unit score of each proposal by each expert, which we shall denote as $P_{ik}$, is as follows

$$P_{ik} = \text{Min}_j [Neg(I(q_j)) \vee P_{ik}(q_j)]$$

In the above $\vee$ indicates the max operation. We first note that this formulation can be implemented on elements drawn from a linear scale as it only involves max, min and negation.

This formulation can be seen as a generalization of a weighted averaging. Linguistically, this formulation is saying that

*if the criteria is important then a proposal should score well on it.*

Essentially this methodology starts off by assuming each project has a score of perfect and then reduces its evaluation by its scoring on each question. However, the amount of this reduction is limited by the importance of the criteria as manifested by the negation. A more detailed discussion of this methodology can be found in Yager (1981).

**Example:** We shall use the previous manifestation to provide an example

| Criteria:   | $Q_1$ | $Q_2$ | $Q_3$ | $Q_4$ | $Q_5$ | $Q_6$ |
|---|---|---|---|---|---|---|
| Importance: | P  | VH | VH | M  | L  | L  |
| Score:      | H  | M  | L  | P  | VH | P  |

In this case

$P_{ik}$ = Min [Neg(P) $\vee$ H, Neg(VH) $\vee$ M, Neg(VH) $\vee$ L, Neg(M) $\vee$ P, Neg(L) $\vee$ VH, Neg(L) $\vee$ P]

$P_{ik}$ = Min [N $\vee$ H, VL $\vee$ M, VL $\vee$ L, M $\vee$ P, H $\vee$ VH, H $\vee$ P]

$P_{ik}$ = Min [H, M, L, P, VH, P]

$P_{ik}$ = L

The essential reason for the low performance of this object is that it performed low on the third criteria which has a very high importance. We note that if we change the importance of the third criteria to low, then the proposal would evaluate to medium.

The essential feature of this approach is that we have obtained a reasonable unit evaluation of each proposal by each expert using an easily manageable linguistic scale. We had no need to use numeric values and force undue precision on the experts.

## 4. Combining Expert's Opinions

As a result of the previous section, we have for each proposal, assuming there are r experts, a collection of evaluations $P_{i1}, P_{i2}, \ldots P_{ir}$ where $P_{ik}$ is the unit evaluation of the $i^{th}$ proposal by the $k^{th}$ expert. In this section, we shall provide a technique for combining the expert's evaluation to obtain an overall evaluation for each proposal, which we shall denote as $P_i$. The technique we shall use is based upon the ordered weighted averaging (OWA) operators introduced by Yager (1988) and extended to the linear environment in Yager (To Appear).

The first step in this process is for the decision maker to provide an aggregation function which we shall denote as Q. This function can be seen as a generalization of the idea of how many experts he feels need to agree on a project for it to be acceptable. In particular for each number i where i runs from 1 to r the decision maker

Yager

.t provide a value Q(i) indicating how satisfied he ould be in selecting a proposal that i of the experts where satisfied with. The values for Q(i) should be drawn from the scale $S = \{S_1, S_2, \ldots S_n\}$ described above.

It should be noted that Q(i) should have certain characteristics to make it rational:

(1) As more experts agree the decision maker's satisfaction or confidence should increase;

$$Q(i) \geq Q(j) \quad i > j$$

(2) If all the experts are satisfied then his satisfaction should be the highest possible;

$$Q(r) = \text{Perfect}$$

A number of special forms for Q are worth noting. [4]:

(1) If the decision maker requires all experts to support a proposal

$$Q(i) = \text{none} \quad \text{for } i < r$$
$$Q(r) = \text{perfect}$$

(2) If the support of just one expert is enough to make a proposal worthy of consideration then

$$Q(i) = \text{perfect for all } i$$

(3) If at least m experts' support is needed for consideration then

$$Q(i) = \text{none} \quad i < m$$
$$Q(i) = \text{perfect} \quad i \geq m.$$

We note that while these examples only use the two extreme values of the scale S this not at all necessary or preferred.

In the following we shall suggest a manifestation of Q that can be said to emulate the usual arithmetic averaging function. In order to define this function, we introduce the operation Int [a] as returning the integer value that is closest to the number a. In the following, we shall let n be the number of points on the scale (the cardinality of S) and r be the number of experts participating. This function which emulates the average is denoted as $Q_A$ and is defined for all $i = 0, 1, \ldots r$ as

$$Q_A(k) = S_{b(k)}$$

where

$$b(k) = \text{Int } [1 + (k * \frac{n-1}{r})].$$

We note that whatever the values of n and r it is always the case that

$$Q_A(0) = S_1$$
$$Q_A(r) = S_n.$$

As an example of this function if r = 3 and n = 7 then

$$b(k) = \text{Int } [1 + (k * \frac{6}{3})] = \text{Int } [1 + 2k]$$

and

$$Q_A(0) = S_1$$
$$Q_A(1) = S_3$$
$$Q_A(2) = S_5$$
$$Q_A(3) = S_7$$

If r = 4 and n = 7 then

$$b(k) = \text{Int } [1 + k * 1.5]$$

and

$$Q_A(0) = S_1$$
$$Q_A(1) = S_3$$
$$Q_A(2) = S_4$$
$$Q_A(3) = S_6$$
$$Q_A(4) = S_7$$

In the case where r = 10 and n=7 then

$$b(k) = \text{Int } [1 + k * \frac{6}{10}]$$

then

$$Q_A(0) = S_1$$
$$Q_A(1) = S_2$$
$$Q_A(2) = S_2$$
$$Q_A(3) = S_3$$
$$Q_A(4) = S_3$$
$$Q_A(5) = S_4$$
$$Q_A(6) = S_5$$
$$Q_A(7) = S_5$$
$$Q_A(8) = S_6$$
$$Q_A(9) = S_6$$
$$Q_A(10) = S_7$$

Having appropriately selected Q we are now in the position to use the ordered weighted averaging (OWA) method (Yager 1988, To Appear) for aggregating the experts opinions. Assume we have r experts, each of which has a unit evaluation for the $i^{th}$ projected denoted $P_{ik}$. The first step in the OWA procedure is to order the $P_{ik}$'s in descending order, thus we shall denote $B_j$ as the $j^{th}$ highest score among the experts unit scores for the project. To find the overall evaluation for the $i^{th}$ project, denoted $P_i$, we calculate

$$P_i = \text{Max}_{j=1, \ldots, r} [Q(j) \wedge B_j].$$

In order to appreciate the workings of this formulation we must realize that $B_j$ can be seen as the worst of the $j^{th}$ top scores. Furthermore Q(j) can be seen as an indication of how important the decision maker feels that the support of at least j experts is. The term $Q(j) \wedge B_j$ can be seen as a weighting of an objects j best scores, $B_j$, and the decision maker requirement that j people support the project, Q(j). The max operation plays a role akin to the summation in the usual numeric averaging procedure. More details on this technique can be found in Yager (1988, To Appear).

**Example:** Assume we have four experts each providing a unit evaluation for project i obtained by the methodology discussed in the previous section.

$$P_{i1} = M$$
$$P_{i2} = H$$
$$P_{i3} = L$$
$$P_{i4} = VH$$

Reordering these scores we get

$$B_1 = VH$$
$$B_2 = H$$

$B_3 = M$

$B_4 = L$.

Furthermore, we shall assume that our decision maker chooses as his aggregation function the average like function, $Q_A$. Then with $r = 4$ and scale cardinality $n = 7$, we obtain

$Q_A(1) = L$     $(S_3)$

$Q_A(2) = M$     $(S_4)$

$Q_A(3) = VH$    $(S_6)$

$Q_A(4) = P$     $(S_7)$

We calculate the overall evaluation as

$P_i = \text{Max} [L \wedge VH, M \wedge H, VH \wedge M, P \wedge L]$

$P_i = \text{Max} [L, M, M, L]$

$P_i = M$

Thus the overall evaluation of this proposal is medium.

Using the methodology suggested thus far we have obtained for each proposal an overall rating $P_i$. These ratings allow us to obtain a comparative evaluation of all the projects without resorting to a numeric scale. The decision maker is now in the position to make his selection of projects to be supported. This decision should combine these comparative linguistic overall ratings with the provided textual material. The process used to make this final decision is not subject to a stringent objective procedure but allows for the introduction of the implicit subjective criteria, such as program agenda, held by the decision maker.

The textual material provided should be used to provide a further distinction amongst the projects.

## 5. CONCLUSION

We have described a methodology to be used in the evaluation of objects which is based upon a non-numeric linguistic scale. The process allows for the multi-criteria evaluation of each object by experts and then a aggregation of this individual experts to obtain an overall object evaluation. This methodology has been suggested as an approach to project funding evaluation where textual material can be used to supplement the selection process. The process can also be used in the information fusion process.